\documentclass[pdflatex,sn-nature]{sn-jnl}

\ifdefined\TemplateGeometry\else
\fi

\usepackage[T1]{fontenc}
\usepackage{lmodern}
\usepackage{graphicx}
\usepackage{amsmath,amssymb}
\usepackage{booktabs}
\usepackage{tabularx}
\usepackage{makecell}
\usepackage{xcolor}

\DeclareUnicodeCharacter{2212}{\ensuremath{-}}
\DeclareUnicodeCharacter{00D7}{\ensuremath{\times}}
\DeclareUnicodeCharacter{00B1}{\ensuremath{\pm}}

\makeatletter\g@addto@macro\UrlBreaks{\do\_}\makeatother

\begin{document}

\title{Evaluating classification and explainability in multimodal petrographic image analysis}

\author[1]{\fnm{Thomas James} \sur{Wood}}
\author*[1,2]{\fnm{Oktay} \sur{Karaku\c{s}}}\email{karakuso@cardiff.ac.uk}
\author[2,3]{\fnm{Can} \sur{Ey\"{u}po\u{g}lu}}

\affil[1]{\orgdiv{Department of Earth and Environmental Sciences}, \orgname{Cardiff University}, \orgaddress{\city{Cardiff}, \country{United Kingdom}}}
\affil[2]{\orgdiv{Department of Computer Science and Informatics}, \orgname{Cardiff University}, \orgaddress{\city{Cardiff}, \country{United Kingdom}}}
\affil[3]{\orgdiv{Department of Computer Engineering}, \orgname{Turkish Air Force Academy, National Defence University}, \orgaddress{\city{Istanbul}, \country{T\"{u}rkiye}}}

\abstract{Automated thin-section classifiers increasingly combine plane-polarised and cross-polarised light (PPL and XPL), two complementary views central to petrographic identification. Classification accuracy alone does not reveal whether model errors reflect mineralogical relationships, how predictions depend on each optical view, or whether attribution maps identify regions that influence model decisions. Here, we examine these complementary aspects of the released LITHOS dual-stream vision transformer using outputs from an earlier evaluation of 105,802 expert-annotated grains across 25 categories. Classification errors frequently followed the label taxonomy: 72.6\% of misclassified plagioclase, microcline and sanidine grains were assigned the generic Feldspar label, while 87.1\% of quartz-family grains received a label within that family, compared with 66.6\% assigned their exact label. Replacing the XPL input with its per-channel spatial means produced a greater reduction in true-class probability than equivalent PPL replacement in 20 of 25 categories, but not in the three quartz categories. Five explainable artificial intelligence (XAI) attribution procedures generated visually distinct maps for identical inputs. In a superpixel-deletion analysis involving 30 grains from 29 thin sections, method-minus-random contrasts ranged from $-0.043$ to $0.057$, with all 41 simultaneous 95\% confidence intervals including zero. Together, these findings show that error structure, modality sensitivity and attribution behaviour capture distinct aspects of model performance that cannot be established from classification accuracy or visual inspection of attribution maps alone.}

\maketitle

\section*{Introduction}

Optical petrography remains a fundamental approach to identifying the minerals, fragments and fossils that constitute rock and soil thin sections. Under a polarising microscope, the same grain is examined in two complementary configurations: plane-polarised light (PPL) reveals colour, pleochroism, relief, grain morphology and cleavage, whereas cross-polarised light (XPL) reveals interference colours, extinction behaviour and twinning \cite{nesse,deer}. Petrographers integrate these observations to identify individual grains, a process that requires considerable expertise, is time-consuming and is difficult to scale. Indeed, annotating the grains in a single large benchmark dataset required approximately one year of full-time expert effort \cite{lithos}. Automating this process has consequently become an established objective in geoscientific image analysis, from early convolutional neural networks applied to carbonate petrography \cite{koeshidayatullah} to curated benchmarks that organise thin-section imagery according to optical modality \cite{deepcarbonate}.

Deep-learning approaches have subsequently diversified in both their architectures and their treatment of complementary optical views. Vision transformers and hybrid convolution--transformer networks have achieved high reported classification accuracy for thin-section lithologies and ceramic fabrics \cite{roder,wei,capriotti}, while explicit fusion of PPL and XPL features has outperformed models relying on either modality alone \cite{ma_fusion}. Evaluation practices have also become more rigorous. In particular, allowing images from the same thin section to appear in both training and test sets can substantially inflate reported accuracy relative to section-wise evaluation \cite{zeng}. The LITHOS benchmark brings these developments together at scale, comprising 105,802 expert-annotated grains across 25 categories, each represented by paired PPL and XPL image patches. It also provides a section-disjoint two-fold evaluation protocol and a released dual-stream vision transformer that integrates modality-specific representations through cross-attention and outperformed single-modality baselines \cite{lithos,kaggle}.

Despite these advances, benchmark accuracy alone provides an incomplete account of petrographic classifier behaviour. Deep neural networks can produce correct predictions by exploiting correlations unrelated to the intended task \cite{geirhos,lapuschkin}. Similarly, object detectors trained on PPL thin-section images have been shown to prioritise grain shape, dominant colour and contrast over the characteristic optical tones of target minerals \cite{morales}. For multimodal petrographic classifiers, two additional questions are particularly relevant. The first concerns the structure of classification errors. Petrographic categories are not equally distinct: LITHOS, like many practical classification schemes, includes specific minerals such as sanidine and microcline alongside a generic feldspar category \cite{lithos}. Consequently, conventional accuracy treats assignment to a broader mineral group as equivalent to a misclassification across unrelated categories, despite their different mineralogical implications. The second question concerns the influence of each optical modality on model predictions. Perturbing or replacing one input provides an operational measure of model sensitivity, but multimodal architectures do not necessarily use their inputs equally or exploit their interactions as intended \cite{hessel,gat}. Petrography provides a particularly informative setting for investigating this issue because the optical information conveyed by PPL and XPL is well established. Whether a learned model exhibits a corresponding pattern of sensitivity remains an empirical question.

Within explainable artificial intelligence (XAI), spatial attribution methods are widely used to investigate the image regions associated with individual model predictions. Grad-CAM assigns class-specific importance to intermediate feature maps using gradients \cite{selvaraju}, attention rollout combines attention information across transformer layers \cite{abnar}, and relevance-propagation approaches integrate attention and gradient information, including in bimodal architectures \cite{chefer_tib,chefer_generic}. However, these procedures capture different aspects of model behaviour, and raw attention weights do not necessarily constitute explanations of model decisions \cite{jain,wiegreffe}. Vision transformers may also produce high-norm artefact tokens that manifest as isolated hotspots in attention maps \cite{darcet}. Nevertheless, thin-section classification studies frequently present Grad-CAM or attention visualisations qualitatively as evidence that models focus on mineralogically or geologically meaningful regions \cite{wei,capriotti}. Findings from other geoscientific applications suggest that such interpretations require caution. On synthetic benchmarks with known ground-truth attributions, commonly used methods produced misleading representations of network behaviour \cite{mamalakis}. Likewise, a quantitative evaluation of Grad-CAM++ explanations for convolutional classifiers of sedimentary structures in core images found that the most accurate network did not necessarily yield the most faithful or stable explanations \cite{abdlmutalib}.

Evaluating attribution quality presents additional methodological challenges. Randomisation-based sanity checks assess whether attribution maps depend on learned model parameters \cite{adebayo}, although their interpretation remains contested \cite{binder}. Perturbation-based metrics instead remove regions ranked as important and measure the resulting change in model output \cite{samek,petsiuk}. Iterative removal of features (IROF) applies this principle using superpixel regions \cite{rieger,achanta}. Although these approaches do not require ground-truth attribution maps, replacing image regions can generate inputs that differ substantially from the training distribution. Consequently, perturbation scores may reflect sensitivity to the replacement procedure as well as the information contained in the removed regions \cite{hooker,rong}. Faithfulness metrics can also exhibit statistical instability and sensitivity to implementation choices \cite{tomsett}, while deletion-based evaluations have struggled to distinguish some vision-transformer attribution methods from random rankings \cite{saco}. Petrographic datasets introduce an additional statistical consideration: grains originating from the same thin section share preparation procedures, acquisition conditions and geological context, and therefore cannot necessarily be treated as independent observations. Uncertainty estimation should account for this clustering at the thin-section level \cite{field_welsh,davison}, while comparisons involving multiple attribution methods and modalities require simultaneous inference \cite{westfall}.

These aspects of model evaluation have rarely been examined together within a single petrographic classification framework. The original LITHOS study established the dataset, model architecture and benchmark performance \cite{lithos}, while an earlier dissertation evaluation of the released model generated the predictions, modality-perturbation responses and attribution maps examined in the present study. Existing petrographic deep-learning studies have generally supplemented classification accuracy with a limited number of qualitative attribution examples \cite{wei,capriotti}. Here, we re-analyse the outputs of the released LITHOS dual-stream transformer, which we did not develop, across four complementary dimensions: (i) the relationship between classification errors and the mineralogical label taxonomy; (ii) category-dependent sensitivity to replacement of either optical modality; (iii) qualitative differences among five attribution procedures applied to identical inputs; and (iv) the deletion-based performance of their spatial importance rankings, for which we newly estimate thin-section-level uncertainty and simultaneous confidence intervals. These dimensions address distinct aspects of model behaviour, and favourable performance under one does not necessarily imply favourable performance under another. Rather than introducing a new classifier, explanation method or perturbation experiment, our objective is to establish what these complementary forms of evidence can, individually and collectively, reveal about the behaviour and interpretability of a multimodal petrographic classifier.

\section*{Results}

\subsection*{Classification errors follow the petrographic label taxonomy}

The model's predictions reproduced the published LITHOS evaluation results. Across the two section-disjoint folds, mean accuracy was 0.623, while macro-averaged recall, precision and F1 were 0.415, 0.519 and 0.432, respectively. These values and their two-fold standard deviations agree, after rounding, with those reported for the released baseline \cite{lithos} (Table~\ref{tab1}). Computing the metrics after pooling predictions from both folds yielded a slightly lower macro-F1 of 0.429. This difference reflects the aggregation procedure rather than a discrepancy in model performance: pooled macro metrics are calculated from combined confusion counts, whereas the published results represent arithmetic means of fold-level estimates.

% Table 1 -- transcribed from Revision3/Manuscript.md (values unchanged)
\begin{table}[tbp]
\centering
\caption{Classification performance of the LITHOS dual-polarisation baseline.}\label{tab1}
\footnotesize
\setlength{\tabcolsep}{3pt}
\begin{tabular*}{\textwidth}{@{\extracolsep{\fill}}lrcccc@{}}
\toprule
Evaluation & Grains & Accuracy & \makecell{Macro\\recall} & \makecell{Macro\\precision} & Macro-F1 \\
\midrule
\multicolumn{6}{@{}l}{Published baseline \cite{lithos}} \\
\quad Two-fold mean $\pm$ s.d. & 105,802 & 0.623 $\pm$ 0.019 & 0.415 $\pm$ 0.006 & 0.519 $\pm$ 0.001 & 0.432 $\pm$ 0.002 \\
\addlinespace
\multicolumn{6}{@{}l}{This study} \\
\quad Fold 1 & 52,792 & 0.6361 & 0.4199 & 0.5199 & 0.4339 \\
\quad Fold 2 & 53,010 & 0.6099 & 0.4110 & 0.5187 & 0.4305 \\
\quad Two-fold mean $\pm$ s.d. & 105,802 & 0.6230 $\pm$ 0.0185 & 0.4155 $\pm$ 0.0063 & 0.5193 $\pm$ 0.0009 & 0.4322 $\pm$ 0.0024 \\
\quad Folds pooled & 105,802 & 0.6230 & 0.4041 & 0.5114 & 0.4286 \\
\botrule
\end{tabular*}
\footnotetext{Macro metrics weight the 25 per-category values equally; s.d.\ is the sample standard deviation of the two fold values. Each fold was evaluated with the model trained on the other fold.}
\end{table}

\begin{figure}[tbp]
\centering
\includegraphics[width=\textwidth]{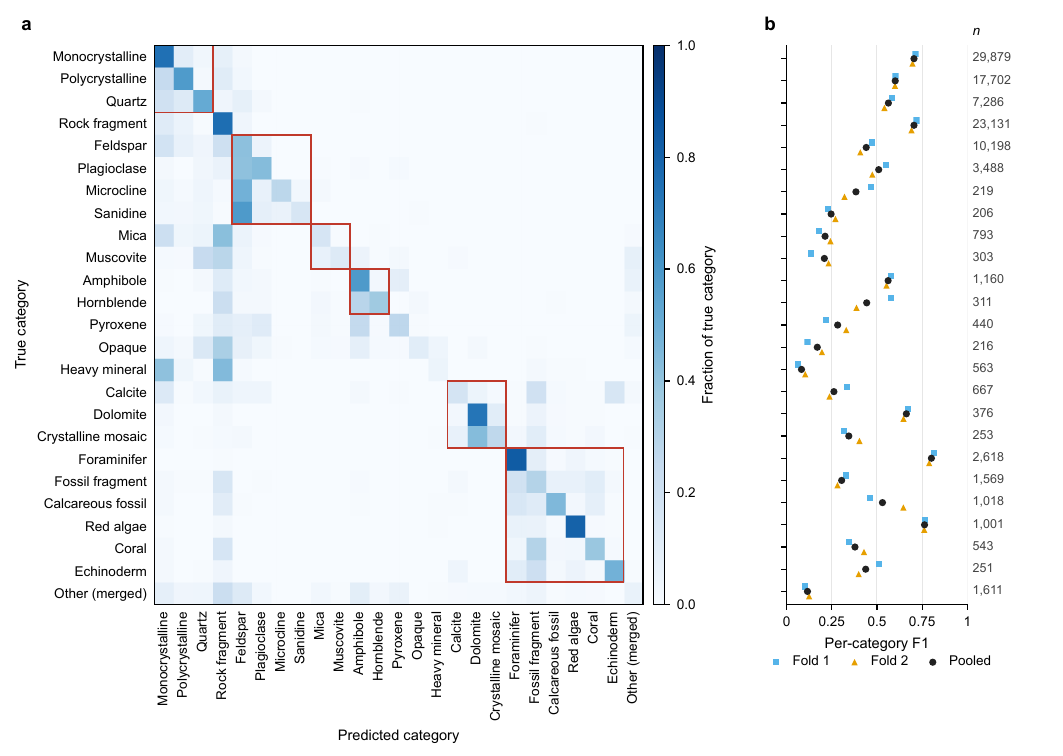}
\caption[Classification performance and error structure of the LITHOS dual-polarisation baseline.]{\textbf{Classification performance and error structure of the LITHOS dual-polarisation baseline.} (\textbf{a}) Row-normalised confusion matrix pooled across both section-disjoint folds (105,802 grains). Each row represents the distribution of predicted categories for grains belonging to a given true category. Categories are arranged by petrographic grouping, with red outlines identifying the quartz family, feldspars, micas, amphiboles, carbonates and biogenic categories discussed in the text. The classifier treats all 25 labels as mutually exclusive. (\textbf{b}) Per-category F1 scores for fold 1 (52,792 grains), fold 2 (53,010 grains) and the pooled predictions, together with the number of grains in each category (\emph{n}). Confidence intervals are not presented because grain-level thin-section identifiers were unavailable for the complete prediction dataset.}\label{fig1}
\end{figure}

The difference between accuracy and macro-F1 reflects the long-tailed category distribution of the dataset \cite{lithos}. Monocrystalline, Rock fragment and Polycrystalline collectively accounted for 66.8\% of grains and achieved pooled F1 scores of 0.705, 0.705 and 0.601, respectively. In contrast, Opaque, Heavy mineral and Other (merged) each had F1 scores below 0.17 (Fig.~\ref{fig1}b). Beyond these aggregate measures, the confusion matrix revealed a structured pattern of classification errors. When categories were arranged according to petrographic relationships, misclassifications were concentrated within groups of related labels rather than distributed uniformly across categories (Fig.~\ref{fig1}a).

The feldspar categories provided the clearest example of this pattern (Table~\ref{tab2}). Among 2,293 misclassified plagioclase, microcline and sanidine grains, 1,664 (72.6\%) were assigned the generic Feldspar label. This accounted for 58.7\% of all sanidine grains, 48.4\% of microcline grains and 41.2\% of plagioclase grains. Reverse misclassifications were considerably less frequent: only 0.12\% and 0.17\% of grains labelled Feldspar were predicted as Microcline or Sanidine, respectively, while 5.8\% were predicted as Plagioclase. Thus, specific mineral labels were assigned less frequently than their occurrence in the reference annotations, with errors predominantly directed towards the generic category. This asymmetry was particularly evident for the potassium feldspars, where precision substantially exceeded recall (0.532 versus 0.160 for sanidine; 0.616 versus 0.279 for microcline), and the number of grains receiving each specific label was less than half the number carrying that reference label. A similar pattern occurred for Hornblende: 91 of its 196 misclassified grains (46.4\%) were assigned to Amphibole, whereas only 2.6\% of Amphibole grains were predicted as Hornblende. Nevertheless, taxonomic relationships did not explain all errors. Muscovite was more frequently misclassified as Rock fragment (28.7\%) or Quartz (24.4\%) than as Mica (7.6\%). Within the feldspar group, exact-label agreement was 41.7\%, increasing to 58.3\% when predictions within the same group were considered. This increase reflects the taxonomic proximity of predictions and reference labels, rather than additional correct classifications under the original 25-category scheme.

% Table 2 -- transcribed from Revision3/Manuscript.md (values unchanged)
\begin{table}[tbp]
\centering
\caption{Selected pooled confusions and the relationship between the labels involved.}\label{tab2}
\footnotesize
\setlength{\tabcolsep}{4pt}
\begin{tabularx}{\textwidth}{@{}llrrr>{\raggedright\arraybackslash}X@{}}
\toprule
\makecell[l]{True\\category} & \makecell[l]{Predicted\\category} & \makecell[r]{Count /\\support} & \makecell[r]{\% of\\true} & \makecell[r]{\%\\reverse} & Relationship of labels \\
\midrule
Sanidine & Feldspar & 121 / 206 & 58.7 & 0.2 & Specific potassium feldspar; generic feldspar label \\
Microcline & Feldspar & 106 / 219 & 48.4 & 0.1 & Specific potassium feldspar; generic feldspar label \\
Plagioclase & Feldspar & 1,437 / 3,488 & 41.2 & 5.8 & Feldspar series; generic feldspar label \\
Hornblende & Amphibole & 91 / 311 & 29.3 & 2.6 & Amphibole-supergroup mineral; group label \\
Muscovite & Mica & 23 / 303 & 7.6 & 0.5 & Mica-group mineral; group label \\
Polycrystalline & Monocrystalline & 4,401 / 17,702 & 24.9 & 10.2 & Quartz textural varieties \\
Quartz & Monocrystalline & 1,436 / 7,286 & 19.7 & 3.4 & Quartz varieties \\
Crystalline mosaic & Dolomite & 110 / 253 & 43.5 & 10.9 & Carbonate texture and carbonate mineral \\
Coral & Fossil fragment & 166 / 543 & 30.6 & 10.8 & Biogenic carbonate categories \\
Pyroxene & Amphibole & 112 / 440 & 25.5 & 9.7 & Different silicate groups; not nested \\
Heavy mineral & Rock fragment & 249 / 563 & 44.2 & 0.1 & Heterogeneous category; dominant class \\
\botrule
\end{tabularx}
\footnotetext{Counts are pooled over both folds; support is the number of grains in the true category. The reverse column gives the percentage of grains of the predicted category that were assigned to the true category of the same row. Label relationships follow the LITHOS documentation \cite{lithos} and standard mineral nomenclature \cite{deer,hawthorne}; the classifier itself treats all 25 labels as mutually exclusive.}
\end{table}

The quartz categories exhibited a similarly structured pattern, although their relationships were primarily textural rather than hierarchical. While 66.6\% of quartz-family grains received their exact label, 87.1\% were assigned to one of the three quartz categories, and 61.4\% of errors involving quartz-family grains remained within that family. The most frequent individual confusion in the entire matrix involved Polycrystalline grains classified as Monocrystalline (24.9\%). Related patterns were observed among carbonate categories, with 43.5\% of Crystalline mosaic grains classified as Dolomite, and among biogenic categories, with 30.6\% of Coral and 21.1\% of Echinoderm grains classified as Fossil fragment. However, several prominent errors did not follow these taxonomic relationships. Pyroxene was classified as Amphibole in 25.5\% of cases, while Heavy mineral grains were frequently assigned to the two most abundant categories, Rock fragment (44.2\%) and Monocrystalline (41.0\%). Similarly, 33.8\% of Opaque grains were classified as Rock fragment. Across the full dataset, Rock fragment and Monocrystalline together accounted for 51.1\% of the 39,891 misclassifications.

\subsection*{Sensitivity to PPL and XPL replacement depends on category}

\begin{figure}[tbp]
\centering
\includegraphics[width=\textwidth]{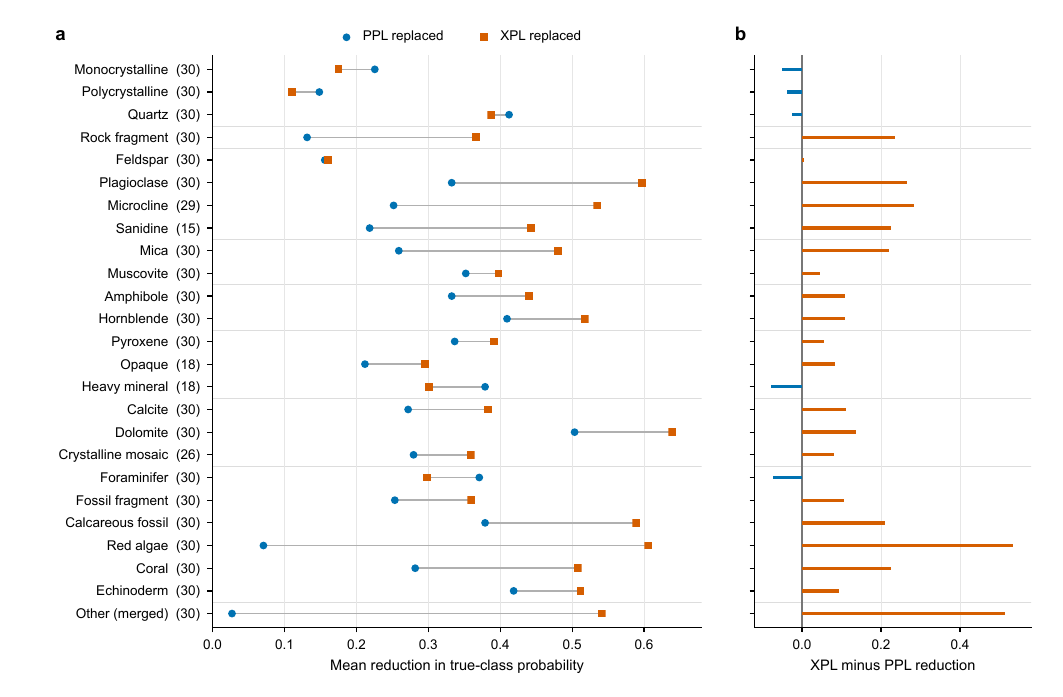}
\caption[Category-dependent sensitivity to replacing the PPL or XPL input.]{\textbf{Category-dependent sensitivity to replacing the PPL or XPL input.} (\textbf{a}) Mean reduction in true-class probability following replacement of the PPL input (circles) or XPL input (squares) with its per-channel spatial means, while leaving the other input unchanged. Results are based on 706 correctly classified fold-2 grains, with category sample sizes shown in parentheses. (\textbf{b}) Difference between the probability reductions following XPL and PPL replacement. Positive values, shown in vermilion, indicate greater sensitivity to XPL replacement and were observed in 20 of 25 categories. Categories follow the ordering in Fig.~\ref{fig1}. Confidence intervals are not shown because only category-level means were retained. Mean replacement preserves average channel values while removing spatial structure and does not isolate the contribution of any individual optical property.}\label{fig2}
\end{figure}

To examine the sensitivity of correct predictions to each optical view, we replaced one input at a time with its per-channel spatial means while keeping the other input unchanged. The resulting reduction in true-class probability was measured for 706 correctly classified grains, with 15--30 grains per category. This intervention preserves the average channel values of the replaced view while removing its spatial structure, including texture, grain boundaries and local contrast. XPL replacement produced a greater mean reduction in true-class probability than PPL replacement in 20 of the 25 categories (Fig.~\ref{fig2}). We describe these categories as XPL-sensitive and the remaining five as PPL-sensitive, strictly in relation to the difference between these two interventions. When weighted by category sample size, the mean probability reductions were 0.281 for PPL replacement and 0.419 for XPL replacement. The corresponding equally weighted category means were 0.281 and 0.416. Category-level contrasts, defined as the XPL-minus-PPL reduction, ranged from $-0.078$ to $0.535$ (Supplementary Table S1).

The five exceptions comprised all three quartz categories (Monocrystalline, $-0.051$; Polycrystalline, $-0.038$; Quartz, $-0.025$), together with Foraminifer ($-0.073$) and Heavy mineral ($-0.078$). Within the feldspar group, the three specific mineral categories exhibited substantially greater sensitivity to XPL than PPL replacement, with contrasts ranging from 0.224 to 0.283. In contrast, the generic Feldspar category showed almost no difference between modalities (0.004), with relatively small probability reductions following either PPL (0.156) or XPL (0.161) replacement. Hornblende and Amphibole showed identical contrasts of 0.108. The largest positive contrasts occurred for Red algae (0.535) and Other (merged) (0.514), whose correct predictions were minimally affected by PPL replacement, with reductions of 0.071 and 0.027, respectively. Dolomite exhibited the greatest sensitivity to both interventions, with probability reductions of 0.503 following PPL replacement and 0.639 following XPL replacement. Because only category-level means were retained, grain- and thin-section-level uncertainty could not be estimated. Consequently, small differences, including those observed among the quartz categories, cannot be considered statistically resolved. Furthermore, these findings apply exclusively to grains that the model classified correctly.

\subsection*{XAI Attribution procedures give contrasting accounts of the same predictions}

\begin{figure}[tbp]
\centering
\includegraphics[width=\textwidth]{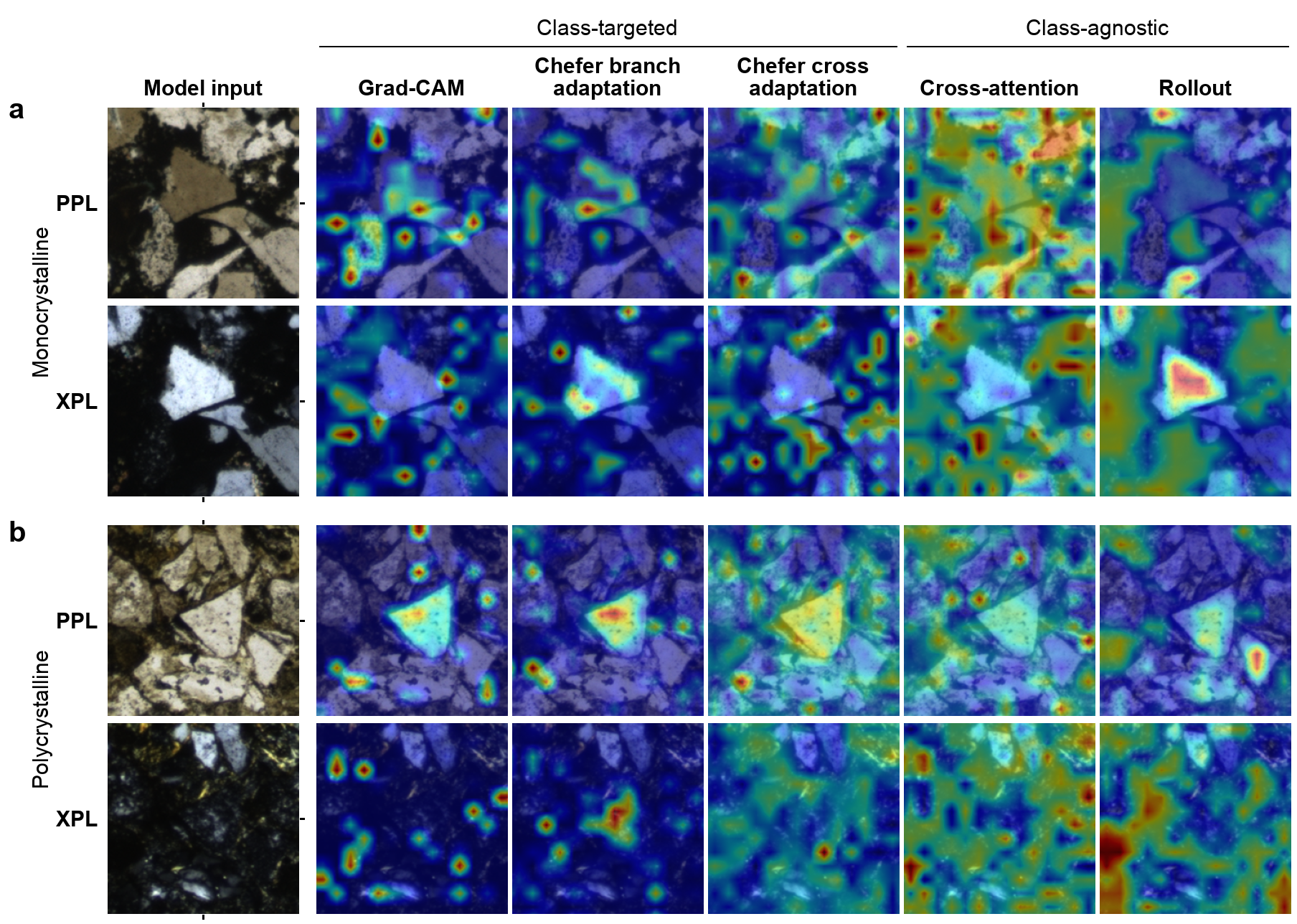}
\caption[Attribution maps from five procedures for two correctly classified grains.]{\textbf{Attribution maps from five procedures for two correctly classified grains.} Rows present the PPL and XPL views of (\textbf{a}) a monocrystalline grain and (\textbf{b}) a polycrystalline grain from fold 2. Both grains were correctly classified, with true-class probabilities of 0.967 and 0.806, respectively. The first column shows the $224 \times 224$ pixel model input; external tick marks indicate the patch centre containing the annotated grain. Subsequent columns show attribution overlays from three class-targeted procedures (Grad-CAM, Chefer branch adaptation and Chefer cross adaptation) and two class-agnostic baselines (decoder cross-attention and attention rollout), following the ordering in Table~\ref{tab3}. Warmer colours indicate higher attribution values. Each map was independently min--max normalised, preventing direct comparison of intensities across panels. Grad-CAM overlays also use a different blending procedure. For cross-attention and Chefer cross attribution, maps over one branch are obtained from the class token of the opposite stream. The examples were selected during the original analysis and were not included in the 30-grain deletion sample; consequently, no deletion scores are available for these grains. Panels were reproduced from previously saved figures without resampling. The maps illustrate differences between procedures and branches but do not establish attribution faithfulness or mineralogical reasoning. Underlying images were obtained from the LITHOS dataset \cite{kaggle} (CC BY-NC-SA 4.0).}\label{fig3}
\end{figure}

Figure~\ref{fig3} compares five attribution procedures applied to two correctly classified grains, alongside the corresponding model inputs. The examples differ in the visibility of the central, annotated grain under each optical configuration. In PPL, the monocrystalline grain (Fig.~\ref{fig3}a) has moderate colour and low contrast relative to its brighter neighbours, whereas in XPL it appears bright against a predominantly dark background. The polycrystalline grain (Fig.~\ref{fig3}b) has a well-defined boundary in PPL, but its central region appears dark and finely mottled in XPL, where its outline is considerably less distinct.

Attribution patterns varied substantially across both procedures and optical branches. For the monocrystalline grain, attention rollout and the Chefer branch adaptation concentrated on the central grain in XPL, whereas their corresponding PPL maps showed no comparable concentration. The opposite pattern was observed for the polycrystalline grain: Grad-CAM and both Chefer adaptations emphasised the central grain in PPL, while their XPL maps were more dispersed or only partially centred. Several recurring characteristics were also apparent. Cross-attention and the Chefer cross adaptation produced relatively diffuse maps with block-like patterns corresponding to the $14 \times 14$ token grid, while Grad-CAM frequently generated isolated token-sized hotspots away from the annotated grain. In both examples, procedures that concentrated attribution on the central grain did so in the optical view where that grain exhibited the strongest contrast with its surroundings.

These examples were selected for illustration during the original analysis and were not included in the deletion sample evaluated below. Accordingly, they have no associated deletion scores. Although they demonstrate that attribution procedures and input branches can produce substantially different visual representations of the same prediction, they do not establish which, if any, of the maps identifies the image regions that influence the model's decision.

\subsection*{Deletion-based evaluation does not resolve differences among procedures}

The spatial rankings generated by each attribution procedure were evaluated using a separate sample of 30 correctly classified grains from 29 thin sections. The sample comprised five grains from each of six categories: Monocrystalline, Polycrystalline, Quartz, Rock fragment, Feldspar and Foraminifer. For each grain and optical branch, superpixels were cumulatively replaced according to their attribution ranking. The resulting normalised area over the true-class probability curve (AOC) was compared with the mean AOC obtained from random segment orderings of the same grain and branch. Higher AOC values indicate an earlier or greater reduction in true-class probability during deletion. The six categories had been selected in the original study and were not representative of the modality-sensitivity patterns identified above. Four categories (Monocrystalline, Polycrystalline, Quartz and Foraminifer) belonged to the five PPL-sensitive categories, Feldspar exhibited nearly balanced sensitivity, and only Rock fragment was clearly XPL-sensitive.

\begin{figure}[tbp]
\centering
\includegraphics[width=\textwidth]{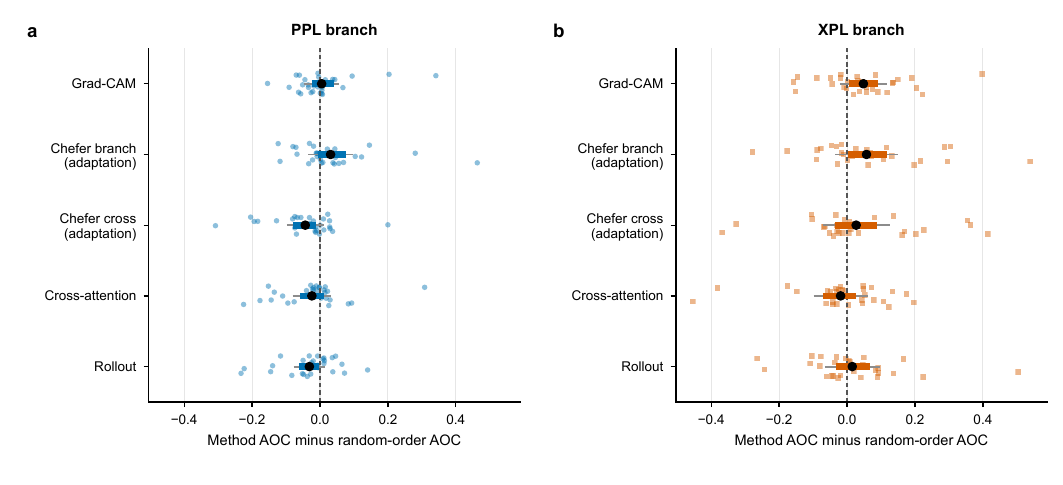}
\caption[Paired deletion contrasts between attribution-ordered and random-ordered superpixel deletion.]{\textbf{Paired deletion contrasts between attribution-ordered and random-ordered superpixel deletion.} Results are shown for (\textbf{a}) the PPL branch and (\textbf{b}) the XPL branch. Each point represents the difference between a grain's AOC under attribution-guided deletion and its mean AOC under random segment ordering on the same branch (30 grains from 29 thin sections). Black circles indicate section-equal means, thick coloured bars indicate pointwise 95\% section-bootstrap confidence intervals, and thin grey bars indicate approximate simultaneous 95\% confidence intervals for the family of 41 contrasts, based on 10,000 bootstrap resamples. The dashed line represents equality with random ordering. All simultaneous confidence intervals include zero. Absolute AOC values are reported in Table~\ref{tab3}.}\label{fig4}
\end{figure}

Random segment ordering yielded section-equal mean AOC values of 0.145 for PPL and 0.183 for XPL (Table~\ref{tab3}), providing a reference for evaluating attribution-guided deletion. On the XPL branch, method-minus-random contrasts were positive for the Chefer branch adaptation (0.057), Grad-CAM (0.048), the Chefer cross adaptation (0.027) and attention rollout (0.015), but negative for cross-attention ($-0.019$). On the PPL branch, the Chefer branch adaptation also yielded a positive contrast (0.031), while Grad-CAM was close to random ordering (0.005). In contrast, rollout ($-0.031$), cross-attention ($-0.024$) and the Chefer cross adaptation ($-0.043$) produced negative point estimates (Fig.~\ref{fig4}). The largest positive XPL contrasts were approximately one-quarter to one-third of the corresponding random-order AOC.

% Table 3 -- transcribed from Revision3/Manuscript.md (values unchanged)
\begin{table}[tbp]
\centering
\caption{Deletion scores and paired differences from random segment order, with equal weighting of thin sections.}\label{tab3}
\footnotesize
\setlength{\tabcolsep}{3pt}
\begin{tabular*}{\textwidth}{@{\extracolsep{\fill}}lrrccr@{}}
\toprule
Procedure and branch & \makecell[r]{Mean\\AOC} & \makecell[r]{Mean minus\\random} & \makecell{Pointwise\\95\% CI} & \makecell{Simultaneous\\95\% CI} & \makecell[r]{Grains above\\random} \\
\midrule
\multicolumn{6}{@{}l}{Random order (reference)} \\
\quad PPL & 0.145 & --- & --- & --- & --- \\
\quad XPL & 0.183 & --- & --- & --- & --- \\
\addlinespace
\multicolumn{6}{@{}l}{Grad-CAM} \\
\quad PPL & 0.150 & 0.005 & [$-$0.024, 0.040] & [$-$0.047, 0.057] & 13/30 \\
\quad XPL & 0.232 & 0.048 & [0.006, 0.092] & [$-$0.021, 0.117] & 20/30 \\
\addlinespace
\multicolumn{6}{@{}l}{Chefer branch adaptation} \\
\quad PPL & 0.176 & 0.031 & [$-$0.006, 0.076] & [$-$0.035, 0.098] & 16/30 \\
\quad XPL & 0.241 & 0.057 & [0.002, 0.119] & [$-$0.035, 0.150] & 19/30 \\
\addlinespace
\multicolumn{6}{@{}l}{Chefer cross adaptation} \\
\quad PPL & 0.101 & $-$0.043 & [$-$0.078, $-$0.011] & [$-$0.097, 0.011] & 9/30 \\
\quad XPL & 0.210 & 0.027 & [$-$0.034, 0.090] & [$-$0.072, 0.126] & 15/30 \\
\addlinespace
\multicolumn{6}{@{}l}{Cross-attention} \\
\quad PPL & 0.121 & $-$0.024 & [$-$0.058, 0.013] & [$-$0.080, 0.033] & 11/30 \\
\quad XPL & 0.164 & $-$0.019 & [$-$0.071, 0.028] & [$-$0.098, 0.061] & 13/30 \\
\addlinespace
\multicolumn{6}{@{}l}{Rollout} \\
\quad PPL & 0.114 & $-$0.031 & [$-$0.061, $-$0.003] & [$-$0.078, 0.015] & 10/30 \\
\quad XPL & 0.199 & 0.015 & [$-$0.033, 0.068] & [$-$0.065, 0.096] & 16/30 \\
\botrule
\end{tabular*}
\footnotetext{Values are means over 29 thin sections, each first averaged over its grains (one section contributes two grains). The random reference is each grain's mean AOC over three shuffled segment orders. Confidence intervals (CIs) are based on 10,000 paired bootstrap resamples of thin sections; simultaneous intervals cover the full family of 41 contrasts (Supplementary Table S2). Differences are computed within grain and branch before averaging and may differ in the third decimal place from the difference of the rounded means. Chefer procedures are architecture-specific adaptations; cross-attention and rollout are class-agnostic structural baselines.}
\end{table}

Four of the ten pointwise confidence intervals in Table~\ref{tab3} excluded zero. These comprised positive XPL contrasts for Grad-CAM and the Chefer branch adaptation, and negative PPL contrasts for the Chefer cross adaptation and attention rollout. Two intervals were particularly close to zero, with a lower bound of 0.0016 for the Chefer branch adaptation and an upper bound of $-0.0032$ for rollout. Recomputing the intervals independently from each half of the bootstrap resamples preserved the direction of all four contrasts. However, their distances from zero were comparable to the Monte Carlo variation between the two halves, which reached 0.0054 across interval endpoints. These pointwise exclusions should therefore be interpreted cautiously. After accounting for the complete family of 41 contrasts, every simultaneous confidence interval included zero. Thus, the analysis did not establish a ranking of attribution procedures, nor did it demonstrate equivalence between any procedure and random ordering.

\begin{figure}[tbp]
\centering
\includegraphics[width=\textwidth]{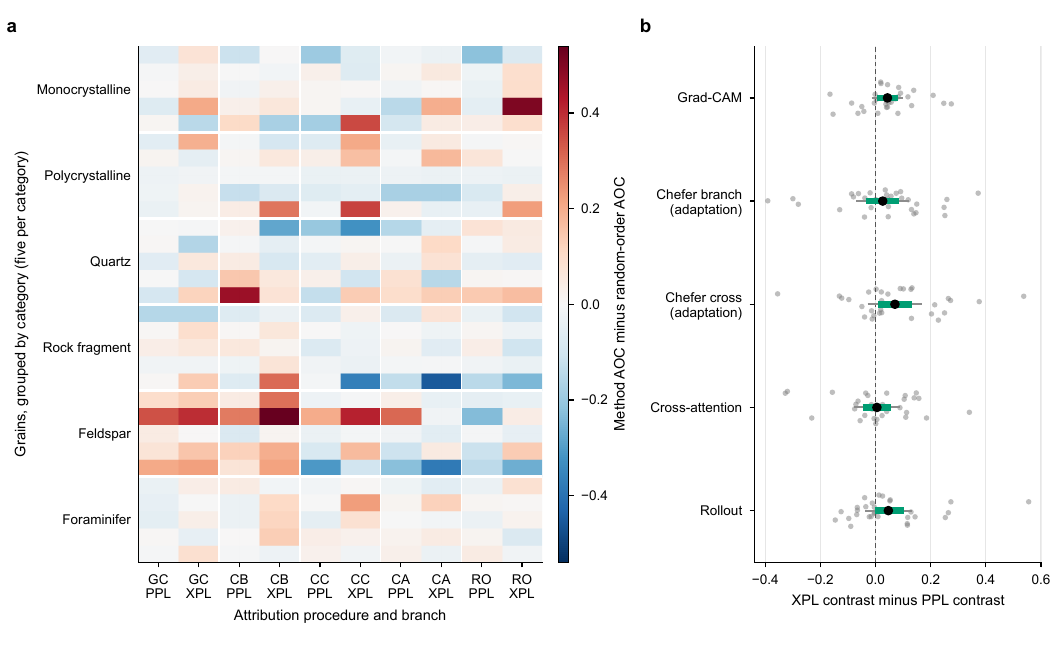}
\caption[Grain-level variability and branch contrasts in the deletion analysis.]{\textbf{Grain-level variability and branch contrasts in the deletion analysis.} (\textbf{a}) Method-minus-random AOC contrasts for individual grains, procedures and optical branches, grouped by category (five grains per category). Colours are scaled symmetrically around zero, with no values truncated. GC, Grad-CAM; CB, Chefer branch adaptation; CC, Chefer cross adaptation; CA, cross-attention; RO, rollout. (\textbf{b}) Within-grain differences between XPL and PPL method-minus-random contrasts for each procedure. Points represent individual grains, black circles indicate section-equal means, thick bars indicate pointwise 95\% confidence intervals, and thin grey bars indicate approximate simultaneous 95\% confidence intervals for the family of 41 contrasts.}\label{fig5}
\end{figure}

Within-grain comparisons between optical branches provided a consistent picture (Fig.~\ref{fig5}b). The XPL-minus-PPL difference in random-adjusted contrasts was positive for all five procedures: 0.070 for the Chefer cross adaptation, 0.047 for rollout, 0.043 for Grad-CAM, 0.026 for the Chefer branch adaptation and 0.005 for cross-attention. For Grad-CAM, the pointwise 95\% confidence interval ranged from 0.007 to 0.080, whereas the simultaneous interval ranged from $-0.015$ to 0.101. All five simultaneous intervals included zero. The random-order AOC was itself 0.039 higher on the XPL branch, although its pointwise interval also included zero ($-0.055$ to 0.129). Accordingly, these data do not establish a clear difference in the overall response of the two branches to progressive deletion.

Mean contrasts masked substantial variability and, in some cases, opposing responses among individual grains (Fig.~\ref{fig5}a). Across all 360 recorded scores, AOC values ranged from $-0.285$ to 0.764. Thirty-nine scores were negative because the perturbed true-class probability exceeded its unperturbed value. For Grad-CAM on the XPL branch, grain-level contrasts ranged from $-0.158$ to 0.398, with 10 of 30 grains performing below random ordering. Nevertheless, the section-equal point estimate was relatively stable under changes in sample composition. It ranged from 0.036 to 0.056 when individual thin sections were omitted and from 0.021 to 0.062 when individual categories were omitted. The grain-equal mean (0.047) was also close to the section-equal mean (0.048). By comparison, the corresponding PPL estimate changed sign when individual thin sections were omitted ($-0.007$ to 0.011), and its median grain-level contrast was negative ($-0.005$).

\section*{Discussion}

Together, these analyses characterise a petrographic classifier that reproduces its published benchmark performance, frequently confuses petrographically related categories, and exhibits greater sensitivity to XPL than PPL replacement in most, but not all, categories. Its XAI attribution procedures also produce visibly different maps, although deletion-based evaluation does not resolve differences among their spatial rankings after simultaneous inference. The value of examining these dimensions together lies in their complementary limitations. Agreement with expert labels does not establish which optical view influences a prediction; sensitivity to replacing an entire view does not identify the influential regions within it; and a visually convincing attribution map does not necessarily rank influential regions more effectively than random ordering. The previously reported divergence between classification accuracy and explanation quality in sedimentary-structure classifiers \cite{abdlmutalib} is therefore relevant here, where the evaluation extends to a dual-polarisation transformer, modality-level and spatial perturbations, and uncertainty estimation at the thin-section level.

The structure of classification errors provides important context for interpreting the model's overall performance. A substantial proportion of errors involving specific feldspar minerals and hornblende consisted of assigning grains to their corresponding generic categories. Although these predictions are taxonomically compatible with the reference labels, they remain incorrect under the original 25-category classification scheme. The higher precision than recall of the specific mineral labels, together with the predominance of errors directed towards generic categories, describes an asymmetry in the distribution of predictions. It does not demonstrate that the model represents uncertainty or deliberately selects a broader taxonomic level, neither of which is explicitly encoded by its flat classification output; prediction calibration was also not evaluated. The considerably larger representation of the generic Feldspar category (10,198 grains, compared with 206--3,488 for the specific feldspar categories) may contribute to this asymmetry. The LITHOS authors attributed confusion among sanidine, microcline and generic feldspar to their similar appearance under polarised light, particularly where diagnostic twinning is difficult to distinguish at patch scale \cite{lithos}. Our findings quantify the pronounced directional imbalance in these errors. Whether the generic Feldspar label was also assigned by annotators when species-level identification was uncertain is not documented in the sources examined. If this occurred, some apparent misclassifications could reflect differences in annotation granularity rather than failures of mineral recognition. Similar considerations apply to the quartz family, where errors frequently involved exchanges between textural varieties, and to carbonate and biogenic categories whose confusions were also associated with shared optical characteristics in the original LITHOS study \cite{lithos}. Nevertheless, conventional macro-averaged metrics assign the same penalty to a sanidine grain classified as Feldspar as to one classified as Opaque.

The distinction between broad-category agreement and exact-label classification is therefore relevant to the practical interpretation of model performance. Agreement within the quartz family reached 87.1\%, while agreement within the feldspar group reached 58.3\%, compared with an exact-label recall of only 16.0\% for sanidine. These findings suggest that predictions retain substantial group-level information even where species-level identification is unreliable, particularly for less frequent categories. Importantly, group-level agreement measures taxonomic proximity and should not be interpreted as a revised classification accuracy. Hierarchy-aware evaluation metrics, or group-level summaries presented alongside exact-label scores, could help determine whether a classifier is suitable for a particular petrographic application, such as group-level modal analysis rather than identification of individual feldspar species. This distinction also draws attention to more substantive errors, including the misclassification of pyroxene as Amphibole and the frequent assignment of Heavy mineral and Opaque grains to the most abundant categories. The original LITHOS study noted the prominence of errors involving the two dominant categories \cite{lithos}; our analysis shows that these categories received slightly more than half of all misclassifications. This suggests that improving performance on underrepresented and heterogeneous categories may be at least as important as resolving distinctions among closely related mineral labels.

Under whole-input replacement, correctly classified grains generally exhibited greater sensitivity to removal of XPL than PPL spatial structure. This pattern is consistent with the importance of interference colour, extinction behaviour and twinning in petrographic identification \cite{nesse}, as well as the reported advantages of dual-polarisation models over single-modality alternatives \cite{lithos,ma_fusion}. However, it describes sensitivity to a specific perturbation and does not establish that XPL is universally more informative or more influential in model predictions. Category-level differences are particularly informative. The three specific feldspar categories were markedly more sensitive to XPL replacement, whereas the generic Feldspar category exhibited little difference between modalities. This is compatible with, but does not demonstrate, a greater dependence on XPL spatial features, potentially including twinning, for species-level predictions. The quartz categories showed the opposite pattern. Polycrystalline quartz is conventionally distinguished by multiple extinction domains visible under XPL \cite{nesse}, yet PPL replacement caused a larger reduction in true-class probability than XPL replacement across all three quartz categories. The polycrystalline example in Fig.~\ref{fig3}b, with a clearly defined boundary in PPL but a less distinct appearance in XPL, illustrates how grain morphology could provide an accessible predictive cue. This interpretation is consistent with evidence that thin-section classifiers may favour shape, colour and contrast over diagnostically relevant optical characteristics \cite{morales}, although a single example cannot establish the mechanism. The contrast between observed model sensitivity and conventional petrographic expectations illustrates information that classification accuracy alone cannot provide. Dolomite offers a complementary example. The LITHOS authors associated its strong performance with rhombohedral morphology and cleavage or twinning visible under XPL \cite{lithos}. Its correctly classified grains were the most sensitive to XPL replacement, but also the most sensitive to PPL replacement, indicating that their predictions were strongly affected by loss of spatial structure in either view.

These interpretations must be considered in relation to the intervention itself. Replacing a view with its per-channel spatial means removes texture and boundaries, disrupts spatial correspondence between the paired inputs, and generates images outside the training distribution \cite{hooker,rong}. Consequently, the measured contrasts reflect sensitivity to this particular replacement operation rather than dependence on any individual optical characteristic. The analysis also includes only correctly classified grains, which may not be representative of all observations, especially for categories with low recall. Complementary approaches, including permutation-based modality importance measures \cite{gat} and methods that isolate cross-modal interactions \cite{hessel}, could provide further insight into how the two input streams contribute to model predictions.

The contrasting XAI attribution maps should be understood as outputs of procedures that characterise different aspects of model behaviour, rather than as interchangeable estimates of a single explanatory truth. Grad-CAM and the Chefer adaptations are class-targeted and incorporate gradient information, whereas cross-attention and attention rollout represent class-agnostic attention patterns. In addition, cross-stream attribution maps describe how the class token of one stream attends to tokens in the other, rather than directly identifying which pixels determine the classification. Differences between these procedures are therefore unsurprising. Some visual characteristics have plausible architectural explanations: block-like patterns correspond to the $14 \times 14$ token grid, while isolated hotspots are compatible with the high-norm artefact tokens previously identified in vision transformers \cite{darcet}, although their origin was not investigated here. Notably, in both illustrated examples, attribution was concentrated on the annotated grain in the view where that grain had the greatest visual contrast with its surroundings. Such concentration is a property of the visualisation, not evidence of attribution faithfulness. It may create an impression of mineralogical relevance even when the maps primarily reflect low-level image characteristics. Accordingly, qualitative attribution maps, as frequently presented in petrographic deep-learning studies \cite{wei,capriotti}, should not be interpreted on their own as evidence that a model has learned mineralogical reasoning.

The deletion-based analysis examined whether differences between attribution maps translated into differences in their ability to rank influential image regions. Grad-CAM and the Chefer branch adaptation, both class-targeted gradient-based approaches, produced the largest positive point estimates on the XPL branch. In contrast, the class-agnostic baselines and the Chefer cross adaptation yielded point estimates at or below random ordering on the PPL branch. This descriptive pattern is compatible with previous findings concerning the potential benefits of gradient information and cross-layer aggregation for transformer attribution \cite{chefer_tib,saco}, as well as concerns about interpreting attention weights as explanations \cite{jain,wiegreffe}. Nevertheless, the present results do not establish a reliable ranking. Method-minus-random contrasts were small relative to the random-order reference, varied substantially between grains, and were not statistically resolved after simultaneous inference. The simultaneous confidence-interval half-widths for the ten method-minus-random contrasts, approximately 0.05--0.10 AOC units, exceeded every corresponding point estimate in magnitude. Difficulties in distinguishing transformer attribution methods from random rankings under cumulative deletion have also been reported elsewhere \cite{saco}. Our findings indicate that, once thin-section-level uncertainty and simultaneous comparisons are considered, a sample of this size cannot reliably distinguish effects of the observed magnitude. The positive XPL estimates for Grad-CAM and the Chefer branch adaptation may therefore motivate targeted confirmation, but they do not establish superiority. Equally, the absence of resolved differences does not demonstrate equivalence to random ordering.

The whole-input and superpixel-deletion analyses share an important operational feature. Both replace selected pixels with the per-channel spatial means of the corresponding preprocessed input. Consequently, replacing every superpixel in a branch produces the same final input as whole-branch replacement for that grain. Each deletion curve therefore describes a progression from the intact input to the whole-branch replacement, with AOC summarising how early the true-class probability decreases along this progression. This operational correspondence does not imply equivalence of evidence. The whole-input analysis was based on 706 grains retained as category-level means, whereas the deletion analysis involved a separately selected sample of 30 grains whose overlap with the former sample could not be established. Neither analysis independently validates the other. Nevertheless, their common endpoint clarifies the distinction between modality sensitivity and spatial attribution quality. A branch may strongly influence a prediction without an attribution procedure identifying its influential regions more effectively than random ordering. Conversely, a small whole-branch effect may limit the detectable benefit of any spatial ranking. Four of the six deletion categories were PPL-sensitive under whole-input replacement, yet the point estimates of within-grain branch contrasts favoured XPL for every attribution procedure. None remained resolved after simultaneous inference, and random-order deletion showed no clear branch asymmetry. A future deletion study stratified across PPL- and XPL-sensitive categories, using the same grains for both interventions, would allow this relationship to be evaluated directly.

These findings have several implications for evaluating multimodal petrographic classifiers. Classification-error analyses should account for the label taxonomy and distinguish errors between related categories from more substantive misidentifications. Modality-perturbation studies should retain grain-level responses and thin-section identifiers to support appropriate uncertainty estimation. Quantitative evaluation of XAI attribution procedures should include a random-order reference matched to each grain and branch, report absolute AOC values alongside random-adjusted contrasts, and account for clustering and multiple comparisons through thin-section-level resampling and simultaneous inference. Qualitative examples would also be more informative if selected from the quantitatively evaluated sample, enabling direct comparison between visual appearance and measured deletion behaviour. Alternative replacement strategies \cite{rong} and attribution-stability analyses \cite{abdlmutalib} could further assess the robustness of the resulting interpretations.

Several limitations constrain the generalisability of these findings. The analyses concern a single released model and rely on experiments conducted before the present study. Both perturbation analyses used one evaluation fold, a single replacement strategy and correctly classified grains only; consequently, attribution behaviour for misclassified grains was not examined. The deletion sample represented six of the 25 categories and comprised grains from 29 thin sections identified through image paths. Some sections may have originated from the same specimens without this relationship being recorded, potentially leading section-level resampling to underestimate uncertainty. Mean-fill deletion can also introduce information through the shapes of replaced regions \cite{rong}, while differences in superpixel area mean that deletion fractions represent proportions of segments rather than pixels. The uncertainty analysis was conditional on the averaged random-order reference because individual random permutations, grain-level deletion curves, superpixel areas and segmentation masks were not retained. Furthermore, the two Chefer procedures were simplified architectural adaptations rather than complete implementations of the original relevance-propagation methods. Finally, the checkpoint identities used in the original experiments were confirmed by the original experimenter but not independently verified, and the perturbation results have not been independently reproduced from the source images and released checkpoints.

Within these limitations, the study demonstrates that a dual-polarisation petrographic transformer can be evaluated along several dimensions beyond conventional classification accuracy. Its errors frequently reflect relationships within the petrographic label taxonomy, its sensitivity to the two optical modalities varies by category and does not always align with conventional petrographic expectations, and its XAI attribution maps differ in ways that the present deletion-based analysis can describe but cannot reliably rank. None of these properties can be inferred directly from the others. Considering them together, with uncertainty estimated at the thin-section level wherever the available records permit, provides a more defensible basis for assessing the suitability of automated petrographic classifiers and their proposed explanations for scientific applications than accuracy scores or illustrative attribution maps alone.

\section*{Methods}

\subsection*{Study design}

This study retrospectively analyses outputs from an earlier evaluation of the released LITHOS multiclass baseline, conducted as part of an MSc dissertation. Chacón and colleagues developed the dataset, expert annotations, fold assignments, dual-stream architecture and trained models \cite{lithos,kaggle}. The original dissertation experiments evaluated these released models and implemented the whole-input replacement, attribution and superpixel-deletion analyses described below. For the present study, classification metrics were recomputed from saved predictions, category-level replacement results were summarised, and paired deletion contrasts were newly estimated with thin-section-level uncertainty using the retained per-grain scores. All numerical inputs were fixed before manuscript preparation. No additional images were classified, no new attribution maps were generated and no model parameters were updated.

The family of comparisons for the deletion analysis was defined retrospectively, after the original point estimates were available. No primary comparison was preregistered, and no independent confirmation sample was used. A parameter-randomisation analysis performed during the original evaluation was excluded because its implementation did not isolate the intended intervention. All recorded deletion scores, including negative values, were retained, and duplicate records of aggregate deletion curves were not treated as independent observations.

\subsection*{Dataset, model and preprocessing}

The LITHOS dataset comprises 105,802 expert-annotated grains extracted from 1,164 high-resolution thin-section images. Each grain is represented by paired $256 \times 256$ pixel PPL and XPL patches centred on the intersection of the annotated grain's major and minor axes, with the reference label corresponding to the central grain \cite{lithos}. Thin sections are assigned to two folds with approximately preserved category distributions and no overlap between folds \cite{lithos}. The folds analysed here contained 52,792 and 53,010 grains, respectively. The released implementation maps 29 original labels to 25 model categories by merging four infrequent labels into the Other category. Classification is performed using a single-label, 25-category output without an explicitly encoded label hierarchy.

The model comprises two ImageNet-pretrained ViT-L/16 encoders \cite{dosovitskiy}, one for each optical modality, whose parameters remain frozen during training \cite{lithos}. Encoder token representations are integrated through decoder layers incorporating cross-attention between the two streams. The decoder outputs are subsequently combined using a single learned scalar weight before classification. Because this weight is a global parameter and each decoder output already incorporates information from both modalities, it cannot be interpreted as a measure of modality dependence for individual grains. During evaluation, images were resized using bicubic interpolation, centre-cropped to $224 \times 224$ pixels, converted to floating-point tensors and normalised using channel means of (0.485, 0.456, 0.406) and standard deviations of (0.229, 0.224, 0.225). These settings follow the torchvision preprocessing preset for the ViT-L/16 IMAGENET1K\_SWAG\_LINEAR\_V1 weights. The original evaluation recorded PyTorch 2.5.0, CUDA execution and a random seed of 0.

Two trained checkpoints were used, with each fold evaluated using the checkpoint trained on the opposite fold. All perturbation and attribution analyses were conducted on fold 2 using the checkpoint trained on fold 1. The original evaluation identified these dual-stream baseline checkpoints as \nolinkurl{LITHOS_BASELINE_FOLD1_25_CLASSES} (trained on fold 1) and \nolinkurl{LITHOS_BASELINE_FOLD2_25_CLASSES} (trained on fold 2). The single-ViT \nolinkurl{best_epoch_f1_fold*.pth} checkpoints were not used. The correspondence between checkpoint identifiers and the original experiments was confirmed by the original experimenter but has not been independently verified.

\subsection*{Classification metrics and error analysis}

Classification metrics were calculated from the saved reference labels and predictions for each evaluation fold. For grains $i = 1, \ldots, N$, with reference label $y_i$ and predicted label $\hat{y}_i$, accuracy was defined as the proportion for which $\hat{y}_i = y_i$. For category $c$, precision, recall and F1 were calculated from the numbers of true positives ($\mathrm{TP}_c$), false positives ($\mathrm{FP}_c$) and false negatives ($\mathrm{FN}_c$):

\begin{equation}
P_c = \frac{\mathrm{TP}_c}{\mathrm{TP}_c + \mathrm{FP}_c}, \qquad
R_c = \frac{\mathrm{TP}_c}{\mathrm{TP}_c + \mathrm{FN}_c}, \qquad
F_c = \frac{2P_cR_c}{P_c + R_c}.
\end{equation}

Macro-averaged precision, recall and F1 were defined as the unweighted means of $P_c$, $R_c$ and $F_c$ across the 25 categories, with undefined ratios assigned a value of zero. These definitions correspond to the macro-averaged scikit-learn metrics used in the released evaluation code. Metrics were calculated separately for each fold and summarised using their arithmetic mean and sample standard deviation, consistent with the published LITHOS results \cite{lithos}. They were also recalculated after pooling predictions from both folds. Probability-based performance metrics were not reproduced because aligned probability vectors were unavailable. Confidence intervals for classification performance were not estimated because grain-level thin-section identifiers were unavailable for the complete prediction dataset.

Classification errors were summarised using the pooled confusion matrix, reporting both counts and percentages relative to the support of each reference category. Petrographic label relationships (Table~\ref{tab2}) were established using the LITHOS documentation \cite{lithos} and standard mineral nomenclature \cite{deer,hawthorne}. For a group of categories $G$, such as the feldspar group (Feldspar, Plagioclase, Microcline and Sanidine) or quartz family (Monocrystalline, Polycrystalline and Quartz), group-level agreement was defined as the proportion of grains with $y_i \in G$ whose predictions also belonged to $G$. These descriptive groupings were not used to modify reference labels or select models. Categories in Fig.~\ref{fig1} were ordered by petrographic relationships for visualisation only.

\subsection*{Whole-input replacement}

Fold-2 grains were matched to their saved predictions, and only those whose highest-probability prediction agreed with the expert label were retained. Up to 30 correctly classified grains per category were sampled without replacement using a seeded NumPy pseudorandom generator (seed 0). This yielded 706 grains across 25 categories, with categories containing fewer eligible grains contributing all available observations (minimum 15).

For grain $i$, let $x_i^{\mathrm{P}}$ and $x_i^{\mathrm{X}}$ denote the preprocessed PPL and XPL inputs, respectively. Let $\pi_i(\cdot,\cdot)$ denote the model's softmax probability for the grain's true class given an input pair, and let $p_i = \pi_i(x_i^{\mathrm{P}},x_i^{\mathrm{X}})$ represent this probability for the intact inputs. For branch $b \in \{\mathrm{P},\mathrm{X}\}$, replacement was performed by setting every pixel in each channel of the preprocessed input to that channel's spatial mean, producing $\tilde{x}_i^{b}$, while leaving the other branch unchanged. The corresponding reductions in true-class probability were defined as

\begin{equation}
\Delta_i^{\mathrm{P}} =
p_i - \pi_i\left(\tilde{x}_i^{\mathrm{P}},x_i^{\mathrm{X}}\right),
\qquad
\Delta_i^{\mathrm{X}} =
p_i - \pi_i\left(x_i^{\mathrm{P}},\tilde{x}_i^{\mathrm{X}}\right).
\end{equation}

Because channel normalisation is an affine transformation, replacement within the normalised tensor is equivalent to filling each channel with its spatial mean before normalisation. For category $c$, modality sensitivity was summarised by the contrast $C_c = \overline{\Delta}_c^{\mathrm{X}} - \overline{\Delta}_c^{\mathrm{P}}$, representing the difference between the mean probability reductions following XPL and PPL replacement. Overall summaries were calculated using both equal category weights and weights proportional to category sample sizes. Only category counts and mean reductions were retained from the original experiments; therefore, grain-level variances, thin-section membership and confidence intervals were unavailable for this analysis.

\subsection*{Attribution procedures}

Five spatial attribution procedures used in explainable artificial intelligence (XAI) were applied separately to each optical branch. Grad-CAM \cite{selvaraju} was implemented using the pytorch-grad-cam library through a model wrapper that held the other modality fixed. The procedure targeted the true-class logit at the output of the first layer normalisation within the final block of the corresponding encoder. The class token was removed, and the remaining 196 tokens were reshaped into a $14 \times 14$ spatial grid.

The Chefer branch adaptation \cite{chefer_tib} replaced each encoder self-attention module with a numerically equivalent explicit computation, yielding a maximum logit difference of $1.12 \times 10^{-5}$ in a parity check. The positive component of the attention--gradient product was averaged across attention heads, propagated through the encoder blocks using a residual relevance matrix, and extracted from the class-token row. The Chefer cross adaptation \cite{chefer_generic} averaged the positive component of gradient-weighted class-token rows from decoder cross-attention across decoder layers. The cross-attention baseline used the same rows without gradient weighting. Attention rollout \cite{abnar} added the identity matrix to head-averaged encoder attention, normalised each row and composed the resulting matrices across layers.

For cross-attention and Chefer cross attribution, the spatial map associated with one branch was obtained from the class token of the opposite stream. Grad-CAM and the Chefer adaptations were targeted to the true class, whereas cross-attention and attention rollout were class-agnostic. All attribution maps were upsampled to $224 \times 224$ pixels and independently min--max normalised. Because deletion rankings depend only on the relative ordering of attribution values, this normalisation did not affect the deletion scores. The two Chefer procedures were adaptations to the LITHOS architecture and were not validated as complete reproductions of the published relevance-propagation algorithms.

\subsection*{Qualitative examples}

During the original evaluation, two illustrative grains per category were sampled using a seeded generator (seed 0) from correctly classified fold-2 observations in the six categories subsequently included in the deletion analysis. All five attribution procedures were applied to the first six examples. Figure~\ref{fig3} presents the first monocrystalline and first polycrystalline examples, whose unperturbed true-class probabilities were 0.967 and 0.806, respectively.

The figure was assembled from previously saved visualisations by copying the image panels at their original resolution ($267 \times 266$ pixels per panel), without resampling, interpolation or colour modification. The original titles and row ordering were removed, and new labels were introduced. Panel-wise comparisons with a second saved visualisation of the same grains yielded Pearson correlations of at least 0.99, confirming that the corresponding panels represented the same attribution maps. Overlays used a jet colour map blended with the input image, although Grad-CAM employed a different blending procedure. Because each map was independently normalised, attribution intensities cannot be compared directly between panels. Neither grain, nor either of their thin sections, was included in the deletion sample, and their membership in the whole-input replacement sample could not be established.

\subsection*{Segmentation and deletion scores}

The deletion sample comprised 30 correctly classified fold-2 grains, with five grains from each of six categories: Monocrystalline, Polycrystalline, Quartz, Rock fragment, Feldspar and Foraminifer. Grains were sampled without replacement within categories using a seeded generator independent of the one used for whole-input replacement. The six categories were selected during the original evaluation and were not intended to represent the overall category distribution. Because the two samples were independently drawn from the same pool of correctly classified grains, overlap was neither required nor excluded, and individual grains could not be linked across the retained datasets.

Each preprocessed branch image was mapped back to the $[0,1]$ intensity range and segmented independently using SLIC superpixels \cite{achanta}, implemented in scikit-image with compactness set to 10. The algorithm initially requested 60 segments; this number was increased successively to 120, 240 and 480 if fewer than 30 segments were produced. Superpixels were ranked by their mean attribution values in descending order, and the resulting sequence was divided into 20 successive groups containing approximately equal numbers of segments.

For each ranking, the corresponding branch was progressively filled with its spatial channel means, one group at a time, while the other branch remained unchanged. This produced 21 probability evaluations per deletion curve, including the intact input. The replacement values were calculated from the original preprocessed input using the same procedure as in the whole-input analysis. Consequently, after the final deletion step, the perturbed branch was identical to $\tilde{x}_i^{b}$. The original evaluation confirmed this correspondence for one grain, with the resulting probabilities agreeing within $5 \times 10^{-4}$.

For a given grain and branch, let $f_k$ denote the cumulative proportion of segments replaced after step $k$, with $f_0 = 0$ and $f_{20} = 1$. Let $p_k$ denote the corresponding true-class probability, where $p_0 = p_i$ is the probability for the intact input. The normalised area over the probability curve (AOC) was defined as

\begin{equation}
\mathrm{AOC} =
1 - \sum_{k=1}^{20}
\left(f_k-f_{k-1}\right)
\frac{p_k+p_{k-1}}{2p_0}.
\end{equation}

This quantity represents the area above the trapezoidally integrated curve of $p_k/p_0$ against the cumulative fraction of deleted segments. Higher AOC values indicate an earlier or greater reduction in true-class probability, while negative values are possible when perturbation increases the probability above its unperturbed value. For each grain and branch, the random-order reference was calculated as the mean AOC from three random permutations of the same superpixels. Only the averaged random reference was retained.

\subsection*{Paired contrasts and section-level uncertainty}

The retained deletion dataset formed a complete grid of 30 grains, two optical branches and six deletion orderings, comprising five attribution procedures and one random-order reference. Thin-section identities were reconstructed from the batch and section components of the image paths, yielding 29 distinct sections, one of which contributed two Foraminifer grains.

For contrast $j$, let $d_{ij}$ denote the paired difference for grain $i$. For example, a method-minus-random contrast was calculated by subtracting the random-reference AOC from the attribution-guided AOC for the same grain and branch. Paired differences were first averaged within each thin section $s$ to obtain $\overline{d}_{sj}$, after which section-level means were weighted equally:

\begin{equation}
\hat{\theta}_j =
\frac{1}{S}\sum_{s=1}^{S}\overline{d}_{sj},
\qquad
\hat{\sigma}_j =
\frac{\mathrm{SD}_s\left(\overline{d}_{sj}\right)}{\sqrt{S}},
\qquad S=29,
\end{equation}

where $\mathrm{SD}_s$ denotes the sample standard deviation across thin sections. Absolute AOC means reported in Table~\ref{tab3} were calculated using the same section-equal weighting. Branch contrasts were obtained by subtracting the PPL method-minus-random difference from the corresponding XPL difference within each grain. Direct comparisons between attribution procedures were similarly calculated from paired scores for the same grain and branch.

The complete comparison family comprised 41 contrasts: 10 method-minus-random comparisons, five branch contrasts, 20 pairwise comparisons between attribution procedures and six unadjusted XPL-minus-PPL differences, including the random-order reference. Several contrasts were linear combinations of others. Their dependence was accounted for through the joint bootstrap distribution.

Uncertainty was estimated using 10,000 bootstrap resamples of the 29 thin sections, sampled with replacement \cite{field_welsh,davison}. Each bootstrap sample was applied jointly to all 41 contrasts, preserving the pairing of observations within grains and thin sections. Pointwise 95\% confidence intervals were calculated from the 2.5th and 97.5th percentiles of the bootstrap estimates $\hat{\theta}_j^{*}$. Approximate simultaneous 95\% confidence intervals were obtained from the bootstrap distribution of the maximum standardised deviation across the comparison family \cite{westfall}:

\begin{equation}
M^{*} =
\max_j
\frac{\left|\hat{\theta}_j^{*}-\hat{\theta}_j\right|}
{\hat{\sigma}_j},
\qquad
\hat{\theta}_j \pm q\hat{\sigma}_j,
\end{equation}

where $q$ denotes the 95th percentile of $M^{*}$, estimated as $q=3.085$. The observed standard errors $\hat{\sigma}_j$ were held fixed rather than recalculated within each bootstrap sample. Accordingly, simultaneous coverage is approximate and relies on the exchangeability of thin sections. No P values were calculated, and no equivalence margin was specified or tested.

As a supplementary check on the influence of comparison-family size, the two largest positive contrasts were evaluated using a Bonferroni correction restricted to those two comparisons. The ratios $\hat{\theta}_j/\hat{\sigma}_j$ were 2.16 for Grad-CAM and 1.91 for the Chefer branch adaptation on the XPL branch, both below the two-sided normal critical value of 2.24.

Monte Carlo precision was assessed by recalculating pointwise confidence intervals independently from the first and second halves of the bootstrap resamples. The largest difference across the 82 interval endpoints was 0.0054. Grain-equal estimates and analyses omitting one thin section, grain or category at a time were used as descriptive sensitivity checks. Statistical analyses were performed using Python 3.12 and NumPy 1.26.4. Figures were generated with Matplotlib 3.9.1, and classification summaries were cross-checked using scikit-learn 1.5.1.

\subsection*{Use of AI-assisted tools}

Large language model tools (OpenAI Codex and Anthropic Claude) assisted with organising analysis records, checking reference metadata, assembling Figure~\ref{fig3} from previously generated image panels, and drafting and revising the manuscript. These tools did not generate experimental data, attribution maps or statistical results. The authors take full responsibility for the manuscript's content.

\backmatter

\bmhead{Data availability}

The LITHOS images, annotations and fold assignments are publicly available through the official Kaggle release (version 1) under a CC BY-NC-SA 4.0 licence \cite{kaggle}. The pretrained models are accessible through the link provided in the LITHOS publication, which specifies the same licence \cite{lithos}. The numerical inputs, analysis outputs, figure source data and Supplementary Tables S1 and S2 supporting the present study are maintained in a private GitHub repository (\url{https://github.com/oktaykarakus/petrographic-xai-evaluation}). The repository will be made publicly accessible upon publication. During peer review, the materials can be provided confidentially to editors and reviewers through an access arrangement agreed with the journal.

\bmhead{Code availability}

The LITHOS model architecture, data loader and preprocessing code are publicly available at \url{https://github.com/BCV-Uniandes/LITHOS} under the MIT licence. The code used for statistical analysis, manuscript figures and supplementary tables, together with the original perturbation and attribution experiment code, is maintained in the study repository described above. It will be released alongside the supporting data and can be made available confidentially to editors and reviewers during peer review.

\bmhead{Acknowledgements}

We thank I. D. Chacón, P. Ruiz Puentes, J. Pearse and P. Arbeláez for making the LITHOS dataset, pretrained models and implementation publicly available.

\bmhead{Funding}

The authors received no specific funding for this work.

\bmhead{Author contributions}

T.J.W. conducted the original experiments, performed the computational analyses and contributed to the initial research investigation. O.K. supervised the research, contributed to its conceptualisation and methodology, and led the development and revision of the manuscript for publication. C.E. contributed to the conceptualisation of the study, evaluation and interpretation of the findings, and critical review of the manuscript. All authors reviewed and approved the final manuscript.

\bmhead{Competing interests}

The authors declare no competing interests.

\bibliography{references}

@book{nesse,
  author    = {Nesse, W. D.},
  title     = {{Introduction to Optical Mineralogy}},
  edition   = {4th},
  publisher = {Oxford Univ. Press},
  year      = {2013}
}

@book{deer,
  author    = {Deer, W. A. and Howie, R. A. and Zussman, J.},
  title     = {{An Introduction to the Rock-Forming Minerals}},
  edition   = {3rd},
  publisher = {Mineralogical Society of Great Britain and Ireland},
  year      = {2013},
  doi       = {10.1180/DHZ},
  url       = {https://doi.org/10.1180/DHZ}
}

@inproceedings{lithos,
  author    = {Chac{\'o}n, I. D. and {Ruiz Puentes}, P. and Pearse, J. and Arbel{\'a}ez, P.},
  title     = {{Towards automated petrography}},
  booktitle = {Advances in Neural Information Processing Systems},
  volume    = {38},
  year      = {2025},
  doi       = {10.52202/085713-3740},
  url       = {https://doi.org/10.52202/085713-3740}
}

@article{koeshidayatullah,
  author  = {Koeshidayatullah, A. and Morsilli, M. and Lehrmann, D. J. and Al-Ramadan, K. and Payne, J. L.},
  title   = {{Fully automated carbonate petrography using deep convolutional neural networks}},
  journal = {Mar. Pet. Geol.},
  volume  = {122},
  pages   = {104687},
  year    = {2020},
  doi     = {10.1016/j.marpetgeo.2020.104687},
  url     = {https://doi.org/10.1016/j.marpetgeo.2020.104687}
}

@article{deepcarbonate,
  author  = {Li, K. and others},
  title   = {{A dataset and benchmark of carbonate thin-section images for deep learning}},
  journal = {Sci. Data},
  volume  = {13},
  pages   = {340},
  year    = {2026},
  doi     = {10.1038/s41597-026-06633-5},
  url     = {https://doi.org/10.1038/s41597-026-06633-5}
}

@article{roder,
  author  = {Roder, M. and others},
  title   = {{Vision transformers as SOTA models for lithological classification of Brazilian pre-salt rocks}},
  journal = {Appl. Comput. Geosci.},
  volume  = {28},
  pages   = {100301},
  year    = {2025},
  doi     = {10.1016/j.acags.2025.100301},
  url     = {https://doi.org/10.1016/j.acags.2025.100301}
}

@article{wei,
  author  = {Wei, P. and others},
  title   = {{A hybrid local-global feature attention network for thin section rock image classification}},
  journal = {Sci. Rep.},
  volume  = {16},
  pages   = {6446},
  year    = {2026},
  doi     = {10.1038/s41598-026-36669-x},
  url     = {https://doi.org/10.1038/s41598-026-36669-x}
}

@article{capriotti,
  author  = {Capriotti, S. and Devoto, A. and Scardapane, S. and Mignardi, S. and Medeghini, L.},
  title   = {{Interpretable classification of Levantine ceramic thin sections via neural networks}},
  journal = {Mach. Learn. Sci. Technol.},
  volume  = {6},
  pages   = {025076},
  year    = {2025},
  doi     = {10.1088/2632-2153/ade6c4},
  url     = {https://doi.org/10.1088/2632-2153/ade6c4}
}

@article{ma_fusion,
  author  = {Ma, W. and Han, T. and Xu, Z. and Lin, P.},
  title   = {{Feature fusion of single and orthogonal polarized rock images for intelligent lithology identification}},
  journal = {AI Civ. Eng.},
  volume  = {4},
  pages   = {5},
  year    = {2025},
  doi     = {10.1007/s43503-025-00049-7},
  url     = {https://doi.org/10.1007/s43503-025-00049-7}
}

@article{zeng,
  author  = {Zeng, T. and others},
  title   = {{Petrographic identification from small-sample thin-section images using a fine-tuned vision-language contrastive learning model}},
  journal = {Minerals},
  volume  = {16},
  pages   = {937},
  year    = {2026},
  doi     = {10.3390/min16090937},
  url     = {https://doi.org/10.3390/min16090937}
}

@misc{kaggle,
  author       = {{Ruiz Puentes}, P.},
  title        = {{LITHOS-DATASET}},
  howpublished = {\emph{Kaggle}, version 1},
  year         = {2025},
  url          = {https://www.kaggle.com/datasets/f1caa92f0ca6bace86fc8fa6d557d3cac4d2cfe545f5aeb7ea2f141ab68d3720}
}

@article{geirhos,
  author  = {Geirhos, R. and others},
  title   = {{Shortcut learning in deep neural networks}},
  journal = {Nat. Mach. Intell.},
  volume  = {2},
  pages   = {665--673},
  year    = {2020},
  doi     = {10.1038/s42256-020-00257-z},
  url     = {https://doi.org/10.1038/s42256-020-00257-z}
}

@article{lapuschkin,
  author  = {Lapuschkin, S. and others},
  title   = {{Unmasking Clever Hans predictors and assessing what machines really learn}},
  journal = {Nat. Commun.},
  volume  = {10},
  pages   = {1096},
  year    = {2019},
  doi     = {10.1038/s41467-019-08987-4},
  url     = {https://doi.org/10.1038/s41467-019-08987-4}
}

@article{morales,
  author  = {Morales, J. and others},
  title   = {{Visual analysis of deep learning semantic segmentation applied to petrographic thin sections}},
  journal = {Sci. Rep.},
  volume  = {15},
  pages   = {14612},
  year    = {2025},
  doi     = {10.1038/s41598-025-99767-2},
  url     = {https://doi.org/10.1038/s41598-025-99767-2}
}

@inproceedings{hessel,
  author    = {Hessel, J. and Lee, L.},
  title     = {{Does my multimodal model learn cross-modal interactions? It's harder to tell than you might think!}},
  booktitle = {Proc. 2020 Conference on Empirical Methods in Natural Language Processing},
  pages     = {861--877},
  year      = {2020},
  doi       = {10.18653/v1/2020.emnlp-main.62},
  url       = {https://doi.org/10.18653/v1/2020.emnlp-main.62}
}

@inproceedings{gat,
  author    = {Gat, I. and Schwartz, I. and Schwing, A.},
  title     = {{Perceptual score: what data modalities does your model perceive?}},
  booktitle = {Advances in Neural Information Processing Systems},
  volume    = {34},
  pages     = {21630--21643},
  year      = {2021},
  url       = {https://proceedings.neurips.cc/paper_files/paper/2021/hash/b51a15f382ac914391a58850ab343b00-Abstract.html}
}

@inproceedings{selvaraju,
  author    = {Selvaraju, R. R. and others},
  title     = {{Grad-CAM: visual explanations from deep networks via gradient-based localization}},
  booktitle = {Proc. IEEE International Conference on Computer Vision},
  pages     = {618--626},
  year      = {2017},
  url       = {https://openaccess.thecvf.com/content_iccv_2017/html/Selvaraju_Grad-CAM_Visual_Explanations_ICCV_2017_paper.html}
}

@inproceedings{abnar,
  author    = {Abnar, S. and Zuidema, W.},
  title     = {{Quantifying attention flow in transformers}},
  booktitle = {Proc. 58th Annual Meeting of the Association for Computational Linguistics},
  pages     = {4190--4197},
  year      = {2020},
  doi       = {10.18653/v1/2020.acl-main.385},
  url       = {https://doi.org/10.18653/v1/2020.acl-main.385}
}

@inproceedings{chefer_tib,
  author    = {Chefer, H. and Gur, S. and Wolf, L.},
  title     = {{Transformer interpretability beyond attention visualization}},
  booktitle = {Proc. IEEE/CVF Conference on Computer Vision and Pattern Recognition},
  pages     = {782--791},
  year      = {2021},
  url       = {https://openaccess.thecvf.com/content/CVPR2021/html/Chefer_Transformer_Interpretability_Beyond_Attention_Visualization_CVPR_2021_paper.html}
}

@inproceedings{chefer_generic,
  author    = {Chefer, H. and Gur, S. and Wolf, L.},
  title     = {{Generic attention-model explainability for interpreting bi-modal and encoder-decoder transformers}},
  booktitle = {Proc. IEEE/CVF International Conference on Computer Vision},
  pages     = {397--406},
  year      = {2021},
  url       = {https://openaccess.thecvf.com/content/ICCV2021/html/Chefer_Generic_Attention-Model_Explainability_for_Interpreting_Bi-Modal_and_Encoder-Decoder_Transformers_ICCV_2021_paper.html}
}

@inproceedings{jain,
  author    = {Jain, S. and Wallace, B. C.},
  title     = {{Attention is not explanation}},
  booktitle = {Proc. 2019 Conference of the North American Chapter of the Association for Computational Linguistics: Human Language Technologies},
  volume    = {1},
  pages     = {3543--3556},
  year      = {2019},
  doi       = {10.18653/v1/N19-1357},
  url       = {https://doi.org/10.18653/v1/N19-1357}
}

@inproceedings{wiegreffe,
  author    = {Wiegreffe, S. and Pinter, Y.},
  title     = {{Attention is not not explanation}},
  booktitle = {Proc. 2019 Conference on Empirical Methods in Natural Language Processing and the 9th International Joint Conference on Natural Language Processing},
  pages     = {11--20},
  year      = {2019},
  doi       = {10.18653/v1/D19-1002},
  url       = {https://doi.org/10.18653/v1/D19-1002}
}

@inproceedings{darcet,
  author    = {Darcet, T. and Oquab, M. and Mairal, J. and Bojanowski, P.},
  title     = {{Vision transformers need registers}},
  booktitle = {International Conference on Learning Representations},
  year      = {2024},
  url       = {https://proceedings.iclr.cc/paper_files/paper/2024/hash/0b408293619f725fd30162af057e531a-Abstract-Conference.html}
}

@article{mamalakis,
  author  = {Mamalakis, A. and Barnes, E. A. and Ebert-Uphoff, I.},
  title   = {{Investigating the fidelity of explainable artificial intelligence methods for applications of convolutional neural networks in geoscience}},
  journal = {Artif. Intell. Earth Syst.},
  volume  = {1},
  pages   = {e220012},
  year    = {2022},
  doi     = {10.1175/AIES-D-22-0012.1},
  url     = {https://doi.org/10.1175/AIES-D-22-0012.1}
}

@article{abdlmutalib,
  author  = {Abdlmutalib, A. J. and Ayranci, K. and Waheed, U. B. and MacEachern, J. A.},
  title   = {{Explainable artificial intelligence using convolutional neural networks for sedimentary structure classification}},
  journal = {Appl. Comput. Geosci.},
  volume  = {31},
  pages   = {100397},
  year    = {2026},
  doi     = {10.1016/j.acags.2026.100397},
  url     = {https://doi.org/10.1016/j.acags.2026.100397}
}

@inproceedings{adebayo,
  author    = {Adebayo, J. and others},
  title     = {{Sanity checks for saliency maps}},
  booktitle = {Advances in Neural Information Processing Systems},
  volume    = {31},
  year      = {2018},
  url       = {https://proceedings.neurips.cc/paper_files/paper/2018/hash/294a8ed24b1ad22ec2e7efea049b8737-Abstract.html}
}

@inproceedings{binder,
  author    = {Binder, A. and others},
  title     = {{Shortcomings of top-down randomization-based sanity checks for evaluations of deep neural network explanations}},
  booktitle = {Proc. IEEE/CVF Conference on Computer Vision and Pattern Recognition},
  pages     = {16143--16152},
  year      = {2023},
  doi       = {10.1109/CVPR52729.2023.01549},
  url       = {https://doi.org/10.1109/CVPR52729.2023.01549}
}

@article{samek,
  author  = {Samek, W. and Binder, A. and Montavon, G. and Lapuschkin, S. and M{\"u}ller, K.-R.},
  title   = {{Evaluating the visualization of what a deep neural network has learned}},
  journal = {IEEE Trans. Neural Netw. Learn. Syst.},
  volume  = {28},
  pages   = {2660--2673},
  year    = {2017},
  doi     = {10.1109/TNNLS.2016.2599820},
  url     = {https://doi.org/10.1109/TNNLS.2016.2599820}
}

@inproceedings{petsiuk,
  author    = {Petsiuk, V. and Das, A. and Saenko, K.},
  title     = {{RISE: randomized input sampling for explanation of black-box models}},
  booktitle = {Proc. British Machine Vision Conference},
  year      = {2018},
  url       = {https://bmva-archive.org.uk/bmvc/2018/contents/papers/1064.pdf}
}

@misc{rieger,
  author       = {Rieger, L. and Hansen, L. K.},
  title        = {{IROF: a low resource evaluation metric for explanation methods}},
  howpublished = {Preprint at \url{https://arxiv.org/abs/2003.08747}},
  year         = {2020}
}

@article{achanta,
  author  = {Achanta, R. and others},
  title   = {{SLIC superpixels compared to state-of-the-art superpixel methods}},
  journal = {IEEE Trans. Pattern Anal. Mach. Intell.},
  volume  = {34},
  pages   = {2274--2282},
  year    = {2012},
  doi     = {10.1109/TPAMI.2012.120},
  url     = {https://doi.org/10.1109/TPAMI.2012.120}
}

@inproceedings{hooker,
  author    = {Hooker, S. and Erhan, D. and Kindermans, P.-J. and Kim, B.},
  title     = {{A benchmark for interpretability methods in deep neural networks}},
  booktitle = {Advances in Neural Information Processing Systems},
  volume    = {32},
  year      = {2019},
  url       = {https://papers.nips.cc/paper_files/paper/2019/hash/fe4b8556000d0f0cae99daa5c5c5a410-Abstract.html}
}

@inproceedings{rong,
  author    = {Rong, Y. and Leemann, T. and Borisov, V. and Kasneci, G. and Kasneci, E.},
  title     = {{A consistent and efficient evaluation strategy for attribution methods}},
  booktitle = {Proc. 39th International Conference on Machine Learning},
  series    = {PMLR},
  volume    = {162},
  pages     = {18770--18795},
  year      = {2022},
  url       = {https://proceedings.mlr.press/v162/rong22a.html}
}

@article{tomsett,
  author  = {Tomsett, R. and Harborne, D. and Chakraborty, S. and Gurram, P. and Preece, A.},
  title   = {{Sanity checks for saliency metrics}},
  journal = {Proc. AAAI Conf. Artif. Intell.},
  volume  = {34},
  pages   = {6021--6029},
  year    = {2020},
  doi     = {10.1609/aaai.v34i04.6064},
  url     = {https://doi.org/10.1609/aaai.v34i04.6064}
}

@inproceedings{saco,
  author    = {Wu, J. and Kang, W. and Tang, H. and Hong, Y. and Yan, Y.},
  title     = {{On the faithfulness of vision transformer explanations}},
  booktitle = {Proc. IEEE/CVF Conference on Computer Vision and Pattern Recognition},
  pages     = {10936--10945},
  year      = {2024},
  doi       = {10.1109/CVPR52733.2024.01040},
  url       = {https://doi.org/10.1109/CVPR52733.2024.01040}
}

@article{field_welsh,
  author  = {Field, C. A. and Welsh, A. H.},
  title   = {{Bootstrapping clustered data}},
  journal = {J. R. Stat. Soc. Ser. B Stat. Methodol.},
  volume  = {69},
  pages   = {369--390},
  year    = {2007},
  doi     = {10.1111/j.1467-9868.2007.00593.x},
  url     = {https://doi.org/10.1111/j.1467-9868.2007.00593.x}
}

@book{davison,
  author    = {Davison, A. C. and Hinkley, D. V.},
  title     = {{Bootstrap Methods and their Application}},
  publisher = {Cambridge Univ. Press},
  year      = {1997},
  doi       = {10.1017/CBO9780511802843},
  url       = {https://doi.org/10.1017/CBO9780511802843}
}

@book{westfall,
  author    = {Westfall, P. H. and Young, S. S.},
  title     = {{Resampling-Based Multiple Testing: Examples and Methods for p-Value Adjustment}},
  publisher = {Wiley},
  year      = {1993}
}

@article{hawthorne,
  author  = {Hawthorne, F. C. and others},
  title   = {{Nomenclature of the amphibole supergroup}},
  journal = {Am. Mineral.},
  volume  = {97},
  pages   = {2031--2048},
  year    = {2012},
  doi     = {10.2138/am.2012.4276},
  url     = {https://doi.org/10.2138/am.2012.4276}
}

@inproceedings{dosovitskiy,
  author    = {Dosovitskiy, A. and others},
  title     = {{An image is worth 16x16 words: transformers for image recognition at scale}},
  booktitle = {International Conference on Learning Representations},
  year      = {2021},
  url       = {https://arxiv.org/abs/2010.11929v2}
}

\bmhead{Supplementary information}

Supplementary Tables S1 and S2, including the detailed category-level results and paired statistical comparisons, are available in the study's GitHub repository (\url{https://github.com/oktaykarakus/petrographic-xai-evaluation}). The repository is currently private; access can be provided confidentially to editors and reviewers during peer review.
\begin{itemize}
	\item \textbf{Supplementary Table S1.} Whole-input replacement results for each of the 25 categories, including sample size, mean reductions in true-class probability following PPL and XPL replacement, and the corresponding modality contrast.
	\item \textbf{Supplementary Table S2.} Complete results for all 41 paired deletion contrasts, including section-equal and grain-equal means, pointwise and simultaneous 95\% confidence intervals, and grain-level sign counts.
\end{itemize}

\end{document}